\documentclass[runningheads]{llncs}
\usepackage{graphicx}
\usepackage{tikz}
\usetikzlibrary{bayesnet}
\usepackage{amsmath}
\usepackage{algorithm} 
\usepackage{algorithmic}
\usepackage{tablefootnote}
\usepackage{url}
\usepackage{xcolor}

\begin{document}
\title{Contrastive Reasons Detection and Clustering from Online Polarized Debates}
\titlerunning{Contrastive Reasons Detection and Clustering}
%
\author{Amine Trabelsi and Osmar R. Za{\"i}ane}
\authorrunning{Trabelsi and Za{\"i}ane}
%
\institute{Department of Computing Science,
University of Alberta
\email{atrabels,zaiane@ualberta.ca}\\}
\maketitle              

\begin{abstract}
This work tackles
the problem of unsupervised modeling  and extraction of the main contrastive sentential reasons conveyed by divergent viewpoints on polarized issues. It proposes a pipeline approach centered around the detection and clustering of phrases, assimilated to argument facets using a novel Phrase Author Interaction Topic-Viewpoint
model.
The evaluation is based on the informativeness, the relevance and the clustering accuracy of extracted reasons. The pipeline approach shows a significant improvement over state-of-the-art methods in contrastive summarization on online debate datasets.
\end{abstract}

\section{Introduction}
\label{intro}
Online debate forums provide a valuable resource for textual discussions about contentious issues. Contentious issues are controversial topics or divisive entities that usually engender opposing stances or viewpoints. Forum users write posts to defend their standpoint using persuasion, reasons or arguments.
Such posts correspond to
what we describe as contentious documents \cite{Trabelsi:14a,Trabelsi:14c,Trabelsi:15}.
An automatic tool that provides a contrasting overview of the main viewpoints and reasons given by opposed sides, debating an issue,  can be useful for journalists and politicians. It provides them with  systematic summaries and drafting elements on argumentation trends.
In this work, given online forum posts about a contentious issue, we study the problems of unsupervised modeling and extraction, in the form of a digest table, of the main contrastive reasons conveyed by divergent viewpoints.
Table \ref{digest-table} presents an example of a targeted solution in the case of the issue of ``Abortion''.
The digest Table \ref{digest-table} is displayed \`a la ProCon.org or Debatepedia websites, where the viewpoints or stances engendered by the issue are separated into two columns. Each cell of a column contains an argument facet label followed by a sentential reason example.
A sentential reason example is one of the infinite linguistic variations used to express a reason. 
For instance, the sentence ``that cluster of cell is not a person'' and the sentential reason ``fetus is not a human" are different realizations of the same reason. 
For convenience, we will also refer to a sentence realizing a reason as a reason.
\textbf{Reasons} in Table \ref{digest-table} are short sentential excerpts, from forum posts, which explicitly or implicitly express premises or arguments supporting a viewpoint. They correspond to any kind of intended persuasion, even if it does not contain clear argument structures \cite{Habernal:17}.
It should make a reader easily infer the viewpoint of the writer.
\textbf{An argument facet} is an abstract concept corresponding to a low level issue or a subject that frequently occurs within arguments in support of a stance or in attacking and rebutting arguments of opposing stance \cite{Misra:15}.
Similar to the concept of reason, many phrases can express the same facet.
Phrases in bold in Table \ref{digest-table}  correspond to \textbf{argument facet labels}, i.e., possible expressions describing argument facets.
Reasons can also be defined as realizations of facets according to a particular viewpoint perspective.
For instance, argument facet 4 in Table \ref{digest-table} frequently occurs within holders of Viewpoint 1 who oppose abortion. It is realized by its associated reason. The same facet is occurring in  Viewpoint 2, in example 9, but it is expressed by a reason rebutting the proposition in example 4.
Thus, reasons associated with divergent viewpoints can share a common argument facet.
Exclusive facets emphasized by one viewpoint's side, much more than the other, may also exist (see example 5 or 8).
Note that in many cases the facet is very similar to the reason or proposition initially put forward by a particular viewpoint side, see examples 2 and 6, 7.
It can also be a general aspect like ``Birth Control'' in example 5. 
 \begin{table*}[t!]
\centering
\scriptsize
\setlength\tabcolsep{2pt}
\caption{\label{digest-table} Contrastive Digest Table for Abortion. }
\begin{tabular}
{|p{0.3cm}p{2cm}p{3.2cm}|p{0.31cm}p{2cm}p{3.2cm}|}
\hline
\multicolumn {3}{|c}{\textit{View 1 \hspace{5 mm} Oppose}}\vline & \multicolumn {3}{c}{\textit{View 2}\hspace{5 mm} Support}\vline\\
\hline \multicolumn {2}{|c}{\bf Argum. facet label}&\bf Reason & \multicolumn {2}{|c}{\bf Argum. facet label}& \bf Reason\\ \hline
1&\bf Fetus is not human& What makes a fetus not human?&6&\bf Fetus is not human& Fetus is not human\\
2&\bf Kill innocent baby& Abortion is killing innocent baby&7&\bf Right to her body&Women have a right to do what they want with their body\\
3&\bf Woman's right to control her body& Does prostitution involves a woman's right to control her body?&8&\bf Girl gets raped and gets pregnant& If a girl gets raped and becomes pregnant does she really want to carry that man's child?\\
4&\bf Give her child up for adoption& Giving a child baby to an adoption agency is an option if a woman isn't able to be a good parent&9&\bf Giving up a child for adoption& Giving the child for adoption can be just as emotionally damaging as having an abortion\\
5&\bf Birth control& Abortion shouldn't be a form of birth control&10&\bf Abortion is not a murder& Abortion is not a murder\\
\hline
\end{tabular}
\vspace{-10pt}
\end{table*}
 
This paper describes the unsupervised extraction of these argument facets' phrases and their exploitation to generate the associated sentential reasons in a contrastive digest table of the issue.
Our first hypothesis is that detecting the main facets in each viewpoint leads to a good extraction of relevant sentences corresponding to reasons.
Our second hypothesis is that leveraging the reply-interactions in online debate helps us cluster the viewpoints and adequately organize the reasons.

We distinguish three common characteristics of online debates, identified also by 
\cite{Hasan:14} and \cite{Boltuvzic:15}, which make the detection and the clustering of argumentative sentences a challenging task. 
First, the unstructured and colloquial nature of used language makes it difficult to detect well-formed arguments. It makes it also noisy containing non-argumentative portions and irrelevant dialogs.
Second, the use of non-assertive speech acts like rhetorical questions to implicitly express a stance or to challenge opposing argumentation, like examples 1,3 and 8 in Table \ref{digest-table}. 
Third, the similarity in words' usage between facet-related opposed arguments leads clustering to errors. Often a post rephrases the opposing side's premise while attacking it (see example 9).
Note that exploiting sentiment analysis solely, like in product reviews, cannot help distinguishing viewpoints. Indeed, Mohammad et al. \cite{Mohammad:17} show that both positive and negative lexicons are used, in contentious text, to express the same stance. Moreover, opinion is not necessarily expressed through polarity sentiment words, like example 6.

In this work, we do not explicitly tackle or specifically model the above-mentioned problems in contentious documents. However, we propose a generic data driven and facet-detection guided approach joined with posts' viewpoint clustering. It leads to extracting meaningful contrastive reasons and avoids running into these problems. 
Our contributions consist of: (1) the conception and deployment of a novel unsupervised generic pipeline framework producing a contrastive digest table of the main sentential reasons expressed in a contentious issue, given raw unlabeled posts from debate forums;
(2) the devising of a novel Phrase Author Interaction Topic Viewpoint model, which jointly processes phrases of different length, instead of just unigrams, and leverages the interaction of authors in online debates.
The evaluation of the proposed pipeline is based on three measures: the informativeness of the digest as a summary, the relevance of extracted sentences as reasons and the accuracy of their viewpoint clustering. The results on different datasets show that our methodology improves significantly over two state-of-the-art methods in terms of documents' summarization, reasons' retrieval and unsupervised contrastive reasons clustering.

\section{Related Work}
\label{relatedWork}
The objective of argument mining is to automatically detect the theoretically grounded argumentative structures within the discourse and their relationships
\cite{Stab:14,Park:14}.
In this work, we are not interested in recovering the argumentative structures but, instead, we aim to discover the underpinning reasons behind people's opinion from online debates. In this section, we briefly describe some of the argument mining work dealing with social media text and present a number of important studies on Topic-Viewpoint Modeling.
The work on online discussions about controversial issues leverages the interactive nature of these discussions.
Habernal and Gurevych \cite{Habernal:17}  consider rebuttal and refutation as possible components of an argument.
Boltu\v{z}i\'{c} and  \v{S}najder \cite{Boltuvzic:14} classify the relationship in a comment-argument pair as an attack (comment attacks the argument), a support or none.
The best performing model of Hasan and Ng's work \cite{Hasan:14} on Reason Classification (RC) exploits the reply information associated with the posts.
Most of the computational argumentation methods, 
are supervised.
Even
the studies focusing on argument identification 
\cite{Swanson:15,Misra:17}
, usually, rely on predefined lists of manually extracted arguments.
As a first step towards unsupervised identification of prominent arguments from online debates, Boltu\v{z}i\'{c}  and  \v{S}najder \cite{Boltuvzic:15} group argumentative statements into clusters assimilated to arguments. However, only selected argumentative sentences are used as input. In this paper, we 
deal with
raw posts containing both argumentative and non-argumentative sentences.

\label{relatedTV}
Topic-Viewpoint models are extensions of Latent Dirichlet Allocation (LDA) \cite{Blei:03} applied to contentious documents. They hypothesize the existence of underlying topic and viewpoint variables that influence the author's word choice when writing about a controversial issue. The viewpoint variable is also called stance, perspective or argument variable in different studies. 
Topic-Viewpoint models are mainly data-driven approaches which reduce the documents into topic-viewpoint dimensions. A Topic-Viewpoint pair t-v is a probability distribution over unigram words. The unigrams with top probabilities characterize the used vocabulary when talking about a specific topic t while expressing a particular viewpoint v at the same time. Several Topic-Viewpoint models of controversial issues exist \cite{Qiu:13,Trabelsi:14b,Trabelsi:16,Thonet:16}.
Little work is done to exploit these models in order to generate sentential digests or summaries of controversial issues instead of just producing distributions over unigram words. Below we introduce the research that is done in this direction. 

Paul et al. \cite{Paul:10} are the first to introduce the problem of contrastive extractive summarization.
They applied their general approach on online surveys and editorials data.
They propose the Topic Aspect Model (TAM) and use its output distributions to compute similarity scores between sentences. Comparative LexRank, a modified LexRank \cite{Erkan:04}, is run on scored sentences to generate the summary.
Recently, Vilares and He \cite{Vilares:17} propose a topic-argument or viewpoint model called the Latent Argument Model\_LEX (LAM\_LEX).
Using LAM\_LEX, they generate a succinct summary of the main viewpoints 
from a parliamentary debates dataset.
The generation consists of ranking the sentences according to a discriminative score for each topic and argument dimension.
It encourages higher ranking of sentences with words exclusively occurring with a particular topic-argument dimension 
which may not be accurate in extracting the contrastive reasons sharing common words.
Both of the studies, cited above, exploit the unigrams output of their topic-viewpoint modeling. 
In this work, we propose a Topic-Viewpoint modeling of phrases of different length, instead of just unigrams.
We believe phrases allow a better representation of the concept of argument facet. They would also lead to extract a more relevant sentence realization of this latter. Moreover, we leverage the interactions of users in online debates for a better contrastive detection of the viewpoints.
\section{Methodology}
\label{methodology}
Our methodology presents a pipeline approach to generate the final digest table of reasons conveyed on a controversial issue. The inputs are raw debate text and the information about the replies. Below we describe the different phases of the pipeline.
\subsection{Phrase Mining Phase}
The inputs of this module are raw posts (documents). We prepare the data by removing identical portions of text in replying posts.
We remove stop and rare words. We consider working with the stemmed version of the words.

The objective of the phrase mining module is to partition the documents into high quality bag-of-phrases instead of bag-of-words. Phrases are of different length, single or multi-words. We follow the steps of El-Kishky et al. \cite{El-Kishky:14}, who propose a phrase extraction procedure for the Phrase-LDA model.
Given the contiguous words of each sentence in a document, the phrase mining algorithm employs a bottom-up agglomerative merging approach. At each iteration, it merges the best pair of collocated candidate phrases if their statistical significance score exceeds a threshold which is set empirically (set according to \cite{El-Kishky:14} implementation).
The significance score depends on the collocation frequency of candidate phrases in the corpus. It measures their number of standard deviation away from the expected occurrence under an independence null hypothesis. The higher the score, the more likely the phrases co-occur more often than by chance.
\subsection{Topic-Viewpoint Modeling Phase}
\begin{figure}
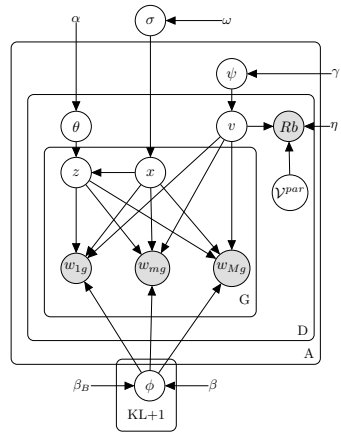

  \centering
  \resizebox{4.5cm}{!} {
  \tikz{ %
    \node[const] (betaB) {$\beta_B$} ; %
    \node[latent, right=of betaB] (phi) {$\phi$} ; %
    \node[const, right=of phi] (beta) {$\beta$} ; %
    \node[obs, above=of phi, xshift=-1.7cm, yshift=1.0cm] (w1) {$w_{1g}$} ; %
    \node[obs, right=of w1] (wi) {$w_{mg}$} ; %
    \node[obs, right=of wi] (wM) {$w_{Mg}$} ; %
    \node[latent, above=of w1,yshift=0.5cm] (z) {$z$} ; %
    \node[latent, above=of z,yshift=-0.65cm] (theta) {$\theta$} ; %
    \node[const, above=of theta,yshift=1.05cm] (alpha) {$\alpha$} ; %
    \node[latent, right=of z] (x) {$x$} ; %
    \node[latent, above=of wM, yshift=1.5cm] (v) {$v$} ; %
    \node[obs, right=of v,xshift=-0.4cm] (r) {$Rb$} ; %
    \node[latent, right=of wM,yshift=1.75cm,xshift=-0.5cm] (V) {$\mathcal{V}^{par}$} ; %
    \node[latent, above=of v,yshift=-0.5cm] (psi) {$\psi$} ; %
    \node[const, right=of psi, xshift=0.95cm] (gamma) {$\gamma$} ; %
    \node[latent, above=of x,yshift=1.8cm] (sigma) {$\sigma$} ; %
    \node[const, right=of sigma,xshift=0.3cm] (omega) {$\omega$} ; %
    \node[const, right=of r,xshift=-0.4cm] (eta) {$\eta$} ; %
    \plate[inner sep=0.25cm, xshift=-0.12cm, yshift=-0.03cm] {plate1} {(w1) (wi) (wM) (z) (x)} {G}; %
    \plate[inner sep=0.25cm, xshift=-0.12cm, yshift=0.12cm] {plate2} {(plate1) (theta) (v) (r) (V)} {D}; %
     \plate[inner sep=0.25cm, xshift=-0.12cm, yshift=0.12cm] {plate3} {(plate2) (psi)} {A}; %
     \plate[inner sep=0.25cm] {plate4} {(phi)} {KL+1}; %
 	\edge {beta} {phi} ;\edge {betaB} {phi} ;
    \edge {phi} {w1} ;  \edge {phi} {wi} ;  \edge {phi} {wM} ; %
    \edge {z} {w1} ; \edge {z} {wi} ;  \edge {z} {wM} ; %
    \edge {x} {w1} ;  \edge {x} {wi} ;  \edge {x} {wM} ;  \edge {x} {z} ;%
    \edge {v} {w1} ;  \edge {v} {wi} ;  \edge {v} {wM} ; %
    \edge {theta} {z} ;
    \edge {v} {r} ;\edge {V} {r} ;\edge {eta} {r} ;
     \edge {psi} {v} ;\edge {gamma} {psi} ;
     \edge {sigma} {x} ;\edge {omega} {sigma} ;
     \edge {alpha} {theta} ;
  }
 }
  \caption{Plate Notation of The PhAITV model}
  \vspace{-10pt}
  \label{figModel}
 
\end{figure}

In this section, we present the Phrase Author Interaction Topic-Viewpoint model (PhAITV). It takes as input the documents, partitioned in high quality phrases of different length, and the information about author-reply interactions in an online debate forum. The objective is to assign a topic and a viewpoint labels to each occurrence of the phrases. This would help to cluster them into Topic-Viewpoint classes.
We assume that $A$ authors participate in a forum debate about a particular issue. Each author $a$ writes $D_a$ posts. Each post $d_a$ is partitioned into $G_{da}$ phrases of different length ($>$=1). Each phrase contain $M_{gda}$ words. Each term $w_{mg}$ in a document belongs to the corpus vocabulary of distinct terms of size $W$. In addition, we assume that we have the information about whether a post replies to a previous post or not.  
Let $K$ be the total number of topics and $L$ be the total number of viewpoints. Let $\theta_{da}$ denote the probability distribution of $K$ topics under a post $d_a$; $\psi_{a}$ be the probability distributions of $L$ viewpoints for an author $a$; $\phi_{kl}$ be the multinomial probability distribution over words associated with a topic $k$ and a viewpoint $l$; and $\phi_{B}$ a multinomial distribution of background words. 
The generative process of a post according to the PhAITV model (see Fig. \ref{figModel}) is the following. An author $a$ chooses a viewpoint $v_{da}$ from the distribution $\psi_{a}$. For each phrase $g_{da}$ in the post, the author samples a binary route variable $x_{gda}$ from a Bernoulli distribution $\sigma$. It indicates whether the phrase is a topical or a background word. Multi-word phrases cannot belong to the background class. If $x_{gda}=0$, she samples the word from $\phi_{B}$. Otherwise, the author, first, draws a topic $z_{gda}$ from $\theta_{da}$, then, samples each word $w_{mg}$ in the phrase from the same $\phi_{z_{gda}v_{da}}$.

Note that, in what follows, we refer to a current post with index $id$ and to a current phrase with index $ig$.
When the current post is a reply to a previous post by a different author, it may contain a rebuttal or it may not. If the reply attacks the previous author then the reply is a rebuttal, and $Rb_{id}$ is set to 1 else if it supports, then the rebuttal takes 0.
We define the \textbf{parent posts} of a current post as all the posts of the author who the current post is replying to. 
Similarly, the \textbf{child posts} of a current post are all the posts replying to the author of the current post. 
We assume that the probability of a rebuttal $Rb_{id}=1$ depends on the degree of opposition between the viewpoint $v_{id}$ of the current post and the viewpoints $\mathcal{V}^{par}_{id}$ of its parent posts as the following:
\begin{equation}
\label{eq1}
p(Rb_{id}=1|v_{id},\mathcal{V}^{par}_{id})=
\frac{\sum\limits^{\mathcal{V}^{par}_{id}}_{l'}\mathbf{I}(v_{id}\neq{l'})+\eta}
{|\mathcal{V}^{par}_{id}|+2\eta},
\end{equation}
where $\mathbf{I}$(condition) equals 1 if the condition is true and $\eta$ a smoothing parameter.

For the inference of the model's parameters, we use the collapsed Gibbs sampling. For all our parameters, we set fixed symmetric Dirichlet priors. According to Fig. \ref{figModel}, the $Rb$ variable is observed. However, the true value of the rebuttal variable is unknown to us. We fix it to 1 to keep the framework purely unsupervised, instead of guiding it by estimating the reply disagreement using methods based on lexicon polarity \cite{Qiu:13}. Setting $Rb=1$ means that all replies of any post are rebuttals attacking all of the parent posts excluding the case when the author replies to his own post. This comes from the observation that the majority of the replies, in the debate forums framework, are intended to attack the previous proposition (see data statistics in Table \ref{stat-table} as an example).  
This setting will affect the viewpoint sampling of the current post. The intuition is that, if an author is replying to a previous post, the algorithm is encouraged to sample a viewpoint which opposes the majority viewpoint of parent posts (Equation \ref{eq1}). Similarly, if the
current post has some child posts, the algorithm is encouraged to sample a viewpoint opposing the children's prevalent stance.
If both parent and child posts exist, the algorithm is encouraged to oppose both, creating some sort of adversarial environment when the prevalent viewpoints of parents and children are opposed. 
The derived sample equation of current post's viewpoint $v_{id}$ given all the previous sampled assignments in the model $\vec{v}_{\neg{id}}$ is:
\begin{multline}
\label{eq2}
p(v_{id}=l|\vec{v}_{\neg{id}},\vec{w},\vec{Rb},\vec{x})\propto
n^{(l)}_{a,\neg{id}}+\gamma
\times
\frac{\displaystyle \prod_{t}^{W_{id}} \prod_{j=0}^{n^{(t)}_{id}-1} n^{(t)}_{l,\neg{id}}+j+\beta}
{\prod_{j=0}^{n_{id}-1}n^{(.)}_{l,\neg{id}}+W\beta+j}\\
\times
p(Rb_{id}=1|v_{id},\mathcal{V}^{par}_{id})
\times
\displaystyle \prod_{c\mid v_{id}\in \mathcal{V}^{par}_{c}} p(Rb_{c}=1|v_{c},\mathcal{V}^{par}_{c}).
\end{multline}
The count $n^{(l)}_{a,\neg{id}}$ is the number of times viewpoint $l$ is assigned to author $a$'s posts excluding the assignment of current post, indicated by $\neg{id}$;
$n^{(t)}_{l,\neg{id}}$ is the number of times term $t$ is assigned to viewpoint $l$ in the corpus excluding assignments in current post;
$n^{(.)}_{l,\neg{id}}$ is the total number of words assigned to $l$;
$W_{id}$ is the set of vocabulary of words in post $id$;
$n^{(t)}_{id}$ is the number of time word $t$ occurs in the post.
The third term of the multiplication in Equation \ref{eq2} corresponds to Equation \ref{eq1} and is applicable when the current post is a reply.
The fourth term of the multiplication takes effect when the current post has child posts. It is a product over each child $c$ according to Equation \ref{eq1}. It computes how much would the children's rebuttal be probable if the value of $v_{id}$ is $l$.

Given the assignment of a viewpoint $v_{id}=l$, we also jointly sample the topic and background values for each phrase $ig$ in post $id$, according to the following:
\begin{multline}
\label{eq3}
p(z_{ig}=k,x_{ig}=1|\vec{z}_{\neg{ig}},\vec{x}_{\neg{ig}},\vec{w},\vec{v})\propto\\
\displaystyle \prod_{j=0}^{M_{ig}}
n^{(1)}_{\neg{ig}}+\omega+j
\times
n^{(k)}_{id,\neg{ig}}+\alpha+j
\times
\frac{n^{(w_{jg})}_{kl,\neg{ig}}+\beta}
{n^{(.)}_{kl,\neg{ig}}+W\beta+j},
\end{multline}
\begin{multline}
\label{eq4}
p(x_{ig}=0|\vec{x}_{\neg{ig}},\vec{w})\propto
\displaystyle \prod_{j=0}^{M_{ig}}
n^{(0)}_{\neg{ig}}+\omega+j
\times
\frac{n^{(w_{jg})}_{0,\neg{ig}}+\beta_B}
{n^{(.)}_{0,\neg{ig}}+W\beta_B+j}.
\end{multline}
Here $n^{(k)}_{id,\neg{ig}}$ is the number of words assigned to topic $k$ 
in post $id$, excluding the words in current phrase $ig$; 
$n^{(1)}_{\neg{ig}}$ and $n^{(0)}_{\neg{ig}}$ correspond to the number of topical and background words in the corpus, respectively; 
$n^{(w_{jg})}_{kl,\neg{ig}}$ and $n^{(w_{jg})}_{0,\neg{ig}}$ correspond to the number of times the word of index $j$ in the phrase $g$ is assigned to topic-viewpoint $kl$ or is assigned as background; $n^{(.)}$s  are summations of last mentioned expressions over all words.

After the convergence of the Gibbs algorithm, each multi-word phrase is assigned a topic $k$ and a viewpoint label $l$. We exploit these assignments to first create clusters $\mathcal{P}_{kl}$s, where each cluster $\mathcal{P}_{kl}$ corresponds to a topic-viewpoint value $kl$. It contains all the phrases that are assigned to $kl$ at least one time. 
Each phrase $phr$
is associated with its total number of assignments.
We note it as $phr.nbAssign$.
Second, we rank the phrases inside each cluster according to their assignment frequencies. 
\subsection{Grouping and Facet Labeling}
\label{groupAndLabel}
The inputs of this module are Topic-Viewpoint clusters, $\mathcal{P}_{kl}$s, $k=1..K$, $l=1..L$, each containing multi-word phrases along with their number of assignments. The outputs are clusters, $\mathcal{A}_{l}$, of sorted phrases corresponding to argument facet labels for each viewpoint $l$ (see Algorithm \ref{algoGroup}).
This phase is based on two assumptions. (1) Grouping constructs agglomerations of lexically related phrases. 
which
can be assimilated to the notion of argument facets.
(2) An argument facet is better expressed with a Verb Expression than a Noun Phrase.
A Verbal Expression (VE) is a sequence of correlated chunks centered around a Verb Phrase chunk \cite{Li:15}.
Algorithm \ref{algoGroup} proposes a second layer of phrase grouping on each of the constructed Topic-Viewpoint cluster $\mathcal{P}_{kl}$ (lines 3-20). It is based on the number of word overlap between stemmed pairs of phrases. The number of groups is not a parameter. First, we compute the number of words overlap between all pairs and sort them in descending order (lines 4-7). Then, while iterating on them, we encourage a pair with overlap to create its own group if both of its phrases are not grouped yet. If it has only one element grouped, the other element joins it. If a pair has no matches, then each non-clustered phrase creates its own group (lines 8-20).

Some of the generated groups may contain small phrases that can be fully contained in longer phrases of the same group. We remove them and add their number of assignments to corresponding phrases. If there is a conflict where two or several phrases can contain the same phrase, then
the one that is a Verbal Expression adds up the number of assignments of the contained phrase. If two or more are VE, then 
the longest phrase, amongst them, adds up the number. 
Otherwise, we prioritize the most frequently assigned phrase (see lines 21-30 in Algorithm \ref{algoGroup}). This procedure helps inflate the number of assignments of Verbal Expression phrases in order to promote them to be solid candidates for the argument facet labeling.
The final step (lines 32-40) consists of collecting the groups pertaining to each Viewpoint, regardless of the topic, and sorting them based on the cumulative number of assignments of their composing phrases. This will create viewpoint clusters, $\mathcal{C}_{l}$s, with groups which are assimilated to argument facets. The labeling consists of choosing one of the phrases as the representative of the group. We simply choose the one with the highest number of assignment to obtain Viewpoint clusters, $\mathcal{A}_{l}$s, of argument facet labels, sorted in the same order of corresponding groups in $\mathcal{C}_{l}$s.
\begin{algorithm}
        \caption{Grouping and Labeling}\label{algoGroup}
        \begin{algorithmic}[1]
            \REQUIRE phrases clusters $\mathcal{P}_{kl}$ for topic $k=1..K$, view $l=1..L$
            \STATE{$\mathcal{G}_{kl} \gets \emptyset$ is the set of groups of phrases to create from $\mathcal{P}_{kl}$}
            \FOR{each phrase cluster $\mathcal{P}_{kl}$}
            \STATE{$\mathcal{Q} \gets$ set of all phrase-pairs from phrases in $\mathcal{P}_{kl}$} 
            \FOR{each phrase-pair $q$ in $\mathcal{Q}$}
            	\STATE{$q.overlap \gets$ number of word intersections in $q$}
            \ENDFOR
            \STATE{Sort pairs in $\mathcal{Q}$ by number of matches in descending order}
             \FOR{each phrase-pair $q$ in $\mathcal{Q}$}
            	\IF{$q.overlap\neq 0$}
                	\IF{$\neg (q.phrase1.grouped) \land \neg(q.phrase2.grouped)  $}
                		\STATE{New group $grp \gets \{q.phrase1\} \cup \{q.phrase2\}$}
                        \STATE{$\mathcal{G}_{kl} \gets \mathcal{G}_{kl} \cup \{grp\} $}
                    \ELSIF{only one phrase of $q$ in existing $grp'$}
                    	\STATE{$grp'\gets grp'\cup \{$non grouped phrase of $q\}$}    
					\ENDIF
                \ELSIF{$\neg q.phrase_{j}.grouped$, $j=1,2$}
                	\STATE{New group $grp \gets \{q.phrase_{j}\}$}
                        \STATE{$\mathcal{G}_{kl} \gets \mathcal{G}_{kl} \cup \{grp\} $}
				\ENDIF
			\ENDFOR
            
            \FOR{each $grp$ in $\mathcal{G}_{kl}$}
            	\STATE{Sort phrases in $grp$ by giving higher ranking to phrases corresponding to: (1) Verbal Expression; (2) longer phrases; (3) frequently assigned phrases}
            	\FOR{each $phr$ in $grp$}
           			\STATE{Find $phr'$ of $grp$ s.t. $phr'.wordSet\subset phr.wordSet$}
           			 \IF{$phr'.nbAssign\neq 0$}
                     	\STATE{$phr.nbAssign \gets phr.nbAssign + phr'.nbAssign$}
                        \STATE{$phr'.nbAssign \gets 0$}
            		 \ENDIF
            	\ENDFOR
            \ENDFOR
\ENDFOR
\STATE{$\mathcal{C}_{l}\gets$ set of all groups belonging to any $\mathcal{G}_{*l}$ of view $l$}
\STATE{$\mathcal{A}_{l}\gets \emptyset$ is the sorted set of all argument facets labels of view $l$}
\FOR{view $l=1$ to $L$}
	\STATE{Sort groups in $\mathcal{C}_{l}$ based on $grp.cumulatifNbAssign$ }
	\FOR{each $grp$ in $\mathcal{C}_{l}$}
    	\STATE{$grp.labelFacet \gets$ phrase with highest $phr.nbAssign$}
        \STATE{$\mathcal{A}_{l} \gets \mathcal{A}_{l}\cup \{grp.labelFacet\}$}
    \ENDFOR
\ENDFOR
\RETURN all clusters $\mathcal{A}_{l}$s of sorted facets' labels for $l=1..L$
\end{algorithmic}
\end{algorithm}
\subsection{Reasons Table Extraction}
The inputs of Extraction of Reasons algorithm are sorted facet labels, $\mathcal{A}_{l}$, for each Viewpoint $l$ (see Algorithm \ref{algoTableExtract}).
Each label phrase is associated with its sentences $\mathcal{S}_{label}$ where it occurs, and where it is assigned a viewpoint $l$.
The target output is the digest table of contrastive reasons $\mathcal{T}$. In order to extract a short sentential reason, given a phrase label, we follow the steps described in Algorithm \ref{algoTableExtract}: 
(1) find, $\mathcal{S}_{label}^{fInters}$, the set of sentences with the most common overlapping words among all the sentences of $\mathcal{S}_{label}$, disregarding the set of words composing the facet label (if the overlap set is empty consider the whole set $\mathcal{S}_{label}$), lines 6-12 in Algorithm \ref{algoTableExtract};
(2) choose the shortest sentence amongst $\mathcal{S}_{label}^{fInters}$  (line 13).
The process is repeated for all sorted facet labels of $\mathcal{A}_{l}$ to fill viewpoint column $\mathcal{T}_l$ for $l=1..L$.
Note that duplicate sentences within a viewpoint column are removed.
If the same sentence occurs in different columns, we only keep the sentence with the label phrase that has the most number of assignments.
Also, we restore stop and rare words of the phrases when rendering them as argument facets.
We choose the most frequent sequence in $\mathcal{S}_{label}$.
\begin{algorithm}
        \caption{Extraction of Reasons Digest Table}\label{algoTableExtract}
\begin{algorithmic}[1]
\REQUIRE all clusters $\mathcal{A}_{l}$s of sorted argument facets' labels for view $l=1..L$;
\STATE{$\mathcal{T}$ is the digest table of contrastive reasons with $\mathcal{T}_{l}$}s columns
\STATE{$\mathcal{T}.columns \gets \emptyset$}
\FOR{view $l=1$ to $L$}
	\STATE{$\mathcal{T}_{l}.cells \gets \emptyset$}
	\FOR{each $label$ in $\mathcal{A}_{l}$}
		\STATE{$\mathcal{S}_{label} \gets$ set of all sentences where $label$ phrase occurs and assigned view $l$}
        \STATE{$fInters\gets$ most frequent set of words overlap among $\mathcal{S}_{label}$ s.t. $fInters\neq label.wordSet$}
        \IF{$fInters\neq \emptyset$}
        	\STATE{$\mathcal{S}_{label}^{fInters} \gets$ subset of $\mathcal{S}_{label}$ containing $fInters$}
        \ELSE
        	\STATE{$\mathcal{S}_{label}^{fInters}\gets \mathcal{S}_{label}$}
        \ENDIF
        \STATE{$sententialReason\gets$ shortest sentence in $\mathcal{S}_{label}^{fInters}$}
        \STATE{$\mathcal{T}_{l}.cells \gets \mathcal{T}_{l}.cells \cup \{cell(label$ + $sententialReason$)\}}  
	\ENDFOR
    \STATE{$\mathcal{T}.columns \gets \mathcal{T}.columns \cup \{\mathcal{T}_{l}\}$}
\ENDFOR
\RETURN{$\mathcal{T}$}
\end{algorithmic}
\end{algorithm}
\section{Experiments and Results}
\label{experiments}
\begin{table}[t!]
\caption{\label{stat-table} Datasets Statistics.
}
\begin{center}
\setlength\tabcolsep{2pt}
\begin{tabular}
{|l|c|c|c|c|c|}
\hline
Forum&\multicolumn {2}{c|}{CreateDebate}&\multicolumn {2}{c|}{4Forums}& Reddit\\
\hline
Dataset&AB& GR& AB& GM& IP\\
\hline 
\# posts& 1876& 1363& 7795 & 6782& 2663\\
\# reason labels& 13& 9& - & -& -\\
\% arg. sent.\tablefootnote{argumentative sentences in the labeled posts}& 20.4& 29.8& - & -& -\\
\% rebuttals& 67.05& 66.61& 77.6& 72.1& -\\
\hline
\end{tabular}
\end{center}
\vspace{-15pt}
\end{table}
\subsection{Datasets}
\label{DS}
We exploit the reasons corpus constructed by \cite{Hasan:14} from the online forum CreateDebate.com, and the Internet Argument corpus containing 4Forums.com datasets \cite{Abbott:16}.
We also scraped a Reddit discussion 
commenting a news article about the March 2018 Gaza clash between Israeli forces and Palestinian protesters \footnote{https://www.reddit.com/r/worldnews/comments/8ah8ys/\\the\_us\_was\_the\_only\_un\_security\_council\_member\_to/}. The constructed dataset does not contain any stance labeling.
We consider 4 other datasets: 
Abortion (AB) and Gay Rights (GR) for CreateDebate, and Abortion and Gay Marriage (GM) for 4Forums. 
Each post in the CreateDebate datasets has a stance label (i.e., support or oppose the issue).
The argumentative sentences of the posts have been labeled in \cite{Hasan:14} with a reason label from a set of predefined reason labels associated with each stance. The reason labels can be assimilated to argument facets or reason types.
Only a subset of the posts, for each dataset, has its sentences annotated with reasons. Table \ref{stat-table} presents some statistics about the data.
Unlike CreateDebate, 4Forums datasets do not contain any labeling of argumentative sentences or their reasons' types. They contain the ground truth stance labels at the author level.
Table \ref{stat-table} reports the percentage of rebuttals as the percentage of replies between authors of opposed stance labels.
The PhAITV model exploits only the text, the author identities and the information about whether a post is a reply or not. 
For evaluation purposes, we leverage the subset of argumentative sentences which is annotated with reasons labels, in CreateDebate, to construct several reference summaries (100) for each dataset. Each reference summary contains a combination of sentences, each from one possible label (13 for Abortion, 9 for Gay Rights). This makes the references exhaustive and reliable resources on which we can build a good recall measure about the informativeness of the digests, produced on CreateDebate datasets. 
\subsection{Experiments Set Up}
We compare the results of our pipeline framework based on \textbf{PhAITV}
to those of two 
studies aiming to produce contrastive summarization in any type of contentious text.
These correspond to Paul et al.'s \cite{Paul:10} 
and Vilares and He's \cite{Vilares:17} works. They are based on Topic-Viewpoint models, \textbf{TAM}, for the first, and \textbf{LAM\_LEX} for the second (see Section \ref{relatedWork}).
Below, we refer to the names of the Topic-Viewpoint methods to describe the whole process that is used to produce the final summary or digest. 
We also compare with a degenerate unigram version of our model, Author Interactive Topic Viewpoint \textbf{AITV} \cite{Trabelsi:18a}. AITV's sentences were generated in a similar way to PhAITV's extraction procedure. The difference is that no grouping is involved and the query of retrieval consists of the top three keywords instead of the phrase.
As a weak baseline, we generate \textbf{random summaries} from the set of possible sentences.
We also create \textbf{correct summaries} from the subset of reason labeled sentences. One correct summary contains all possible reason types of argumentative sentences for a particular issue.
Moreover, we compare with another degenerate version of our model \textbf{PhAITV\textsubscript{view}} which assumes the true values of the posts' viewpoints are given.
Note that the objective here is to assess the final output of the framework. Separately evaluating the performance of the Topic Viewpoint model in terms of document clustering
has shown satisfiable results. We do not report it here for lack of space.

We try different combinations of the PhAITV's hyperparameters and use the combination which gives a satisfying overall performance. 
PhAITV hyperparameters are set as follows: $\alpha=0.1$; $\beta=1$; $\gamma=1$; $\beta_B=0.1$; $\eta=0.01$; $\omega=10$; Gibbs Sampling iterations is $1500$; number of viewpoints $L$ is $2$.
We try a different number of topics $K$ for each Topic-Viewpoint model used in the evaluation. The reported results are on the best number of topics found when measuring the Normalized PMI coherence \cite{Bouma:09} on the Topic-Viewpoint clusters of words.
The values of $K$ are $30$,$10$,$10$ and $50$ for PhAITV, LAM\_LEX, TAM and AITV, respectively.
Other parameters of the methods used in the comparison are set to their default values.
All the models generate their top 15 sentences for Abortion and their 10 best sentences for Gay Rights and Israel-Palestine datasets.
\subsection{Evaluating Argument Facets Detection}
The objective is to verify our assumption that the pipeline process, up to the Grouping and Labeling module, produces phrases that can be assimilated to argument facets' labels.
We evaluate a total of 60 top distinct phrases produced after 5 runs on Abortion (4Forums) and Gay Rights (CreateDebate). We ask two annotators acquainted with the issues, and familiar with the definition of argument facet (Section \ref{intro}), to give a score of 0 to a phrase that does not correspond to an argument facet, a score of 1 to a somewhat a facet, and a score of 2 to a clear facet label. 
Annotator are later asked to find consensus on phrases labeled differently. 
The average scores, of final annotation, on Abortion and Gay Rights are \textbf{1.45} and \textbf{1.44}, respectively.
The percentages of phrases that are not argument facets are \textbf{12.9\%} (AB) and \textbf{17.4\%} (GR). The percentages of clear argument facets labels are \textbf{58.06\%} (AB) and \textbf{62.06\%} (GR).
These numbers validate our assumption that the pipeline succeeds, to a satisfiable degree, in extracting argument facets labels. 
\subsection{Evaluating Informativeness}
\begin{table}[t!]
\begin{center}
\setlength\tabcolsep{4pt}
\caption{\label{rouge-table} Averages of ROUGE Measures (in \%, stemming and stop words removal applied) on Abortion and Gay Rights of CreateDebate. Bold denotes best values, notwithstanding Correct Summaries.}
\begin{tabular}
{lcc|cc}
\hline
& \multicolumn {2}{c|}{Abortion}& \multicolumn {2}{c}{Gay Rights} \\
\hline
&R2-R& R2-FM&R2-R& R2-FM\\
\hline
Rand Summ.& 1.0& 1.0& 0.7& 0.8\\
AITV& 3.0& 2.8& \textbf{2.7}& \textbf{2.8}\\
TAM& 1.8& 2.1& 2.0& 2.4\\
LAM\_LEX& 1.5& 1.0& 1.1& 0.9\\
\hline
PhAITV& \textbf{4.5}& \textbf{4.6}& \textbf{2.7}& \textbf{2.8}\\
\hline
Correct Summ.& 5.8& 5.4& 3.0& 2.9\\
\hline
\end{tabular}
\end{center}
\vspace{-25pt}
\end{table}

We re-frame the problem of creating a contrastive digest table into a summary problem.
The concatenation of all extracted sentential reasons of the digest is considered as a candidate summary. 
The construction of reference summaries, using annotated reasons of CreateDebate datasets, is explained in Section \ref{DS}. The length of the candidate summaries is proportional to that of the references. 
Reference summaries on 4Forums or Reddit datasets can not be constructed because no annotation, of reasons and their types, exists.
We assess all methods, on CreateDebate, using automatic summary evaluation metric ROUGE \cite{Lin:04}.
We report the results of Rouge-2's Recall (R-2 R) and F-Measure (R-2 FM).
Rouge-2 captures the similarities between sequences of bigrams in references and candidates.  
The higher the measure, the better the summary.
All reported ROUGE-2 values are computed after applying stemming and stop words removal on reference and candidate summaries. 
This procedure may also explain the relatively small values of reported ROUGE-2 measures in Table \ref{rouge-table}, compared to those usually computed when stop words are not removed. The existence of stop words in candidate and references sentences increases the overlap, and hence the ROUGE measures' values in general. Applying stemming and stop words removal was based on some  preliminary tests that we conducted on our dataset. The tests showed that two candidate summaries containing different numbers of valid reasons, would have a statistically significant difference in their ROUGE-2 values when stemming and stop words removal applied.

Table \ref{rouge-table} contains the averaged results on 10 generated summaries on Abortion and Gay Rights, respectively. LAM\_LEX performs poorly in this task (close to Random summaries) for both datasets. PhAITV performs significantly better than TAM on Abortion, and slightly better on Gay Rights. Moreover, PhAITV shows significant improvement over its degenerate unigram version AITV on Abortion. This shows that phrase modeling and grouping can play a role in extracting more diverse and informative phrases. AITV beats its similar unigram-based summaries on both datasets. This means that the proposed pipeline is effective in terms of summarization even without the phrase modeling. In addition, PhAITV's ROUGE measures on Gay Rights are very similar to those of the correct summaries (Table \ref{rouge-table}). 
Examples of the final outputs produced by PhAITV framework and the two contenders on Abortion is presented in Table \ref{sample-output}.
The example digests produce proportional results to the median results reported in Table \ref{RelAcc-table}.
We notice that PhAITV's digest produces different types of reasons from diverse argument facets, like putting child up for adoption, life begins at conception, and mother's life in danger. However, such informativeness 
is lacking on both digests of LAM\_LEX and TAM. 
Instead, we remark the recurrence of subjects like killing or taking human life in TAM's digest.
\subsection{Evaluating Relevance and Clustering}
\begin{table*}[t!]
\begin{center}
\setlength\tabcolsep{1.3pt}
\scriptsize
\caption{\label{RelAcc-table} Median values of Relevance Rate (Rel), NPV and Clustering Accuracy Percentages on CreateDebate, FourForums and Reddit Datasets. Bold denotes best results, notwithstanding PhAITV\textsubscript{view}.}
\begin{tabular}
{lccc|ccc|ccc|ccc|ccc}
\hline
&\multicolumn {6}{c|}{CreateDebate}& \multicolumn {6}{c|}{4Forums}& \multicolumn {3}{c}{Reddit}\\
\hline
& \multicolumn {3}{c|}{Abortion}& \multicolumn {3}{c|}{Gay Rights}& \multicolumn {3}{c|}{Abortion}& \multicolumn {3}{c}{Gay Marriage}& \multicolumn {3}{|c}{Isr/Pal}\\
\hline
& Rel& NPV& Acc.&Rel& NPV& Acc.& Rel& NPV& Acc.& Rel& NPV& Acc.& Rel& NPV& Acc.\\
\hline
AITV& 0.66& 58.33& 59.09& 0.5& \textbf{75.0}& 66.66& 0.66& 66.66& 71.42& 0.5& 50.0& 66.66& 0.6&	55.55& 60.00\\
TAM& 0.53& 50.00& 46.42& 0.5& 50.0& 42.85& 0.33& 37.50& 66.66& 0.3& 50.0& 33.33& 0.3& 66.66& 50.00\\
LAM\_LEX& 0.40& 50.00& 64.44& 0.5& 50.0& 50.00& 0.46& 37.50& 46.60& 0.5& 50.0& 50.00& 0.3& 25.00& 33.33\\
\hline
PhAITV& \textbf{0.93}& \textbf{75.00}& \textbf{73.62}& \textbf{0.8}& \textbf{75.0}& \textbf{75.00}& \textbf{0.80}& \textbf{69.44}& \textbf{71.79}& \textbf{0.7}& \textbf{80.0}& \textbf{71.42}& \textbf{0.9}& \textbf{75.00}& \textbf{77.77}\\
\hline
PhAITV\textsubscript{view}& 0.93& 87.50& 83.33& 0.9& 100& 100& 0.80& 83.33& 81.81& 0.9& 100& 100& -& -& -\\
\hline
\end{tabular}
\end{center}
\vspace{-10pt}
\end{table*}
\begin{table*}[t!]
\begin{center}
\setlength\tabcolsep{1.5pt}
\scriptsize
\caption{\label{sample-output} Sample Digest Tables Output of sentential reasons produced by the frameworks based on PhAITV, LAM\_LEX and TAM when using Abortion dataset from CreateDebate. Sentences are labeled according to their stances as the following: (+) reason for abortion; (-) reason against abortion; and (0) irrelevant.}
\begin{tabular}
{|p{0.4cm}p{5.5cm}|p{0.4cm}p{5.5cm}|}
\hline
\multicolumn {4}{|c|}{PhAITV + Grouping + Extraction}\\
\hline
& Viewpoint 1& & Viewpoint 2\\
\hline
(\textbf{-})& \textcolor{red}{If a mother or a couple does not want a child there is always the option of putting the child up for adoption.}
& (\textbf{+})& \textcolor{blue}{The fetus before it can survive outside of the mother's womb is not a person.}\\
(\textbf{-})& \textcolor{red}{I believe life begins at conception and I have based this on biological and scientific knowledge.}& (\textbf{+})& \textcolor{blue}{Giving up a child for adoption can be just as emotionally damaging as having an abortion.}\\
(\textbf{-})& \textcolor{red}{God is the creator of life and when you kill unborn babies you are destroying his creations.}& (\textbf{+})& \textcolor{blue}{you will have to also admit that by definition; abortion is not murder.}\\
(\textbf{-})& \textcolor{red}{I only support abortion if the mothers life is in danger and if the fetus is young.}& (\textbf{-})& \textcolor{red}{No abortion is wrong.}\\
(0)& The issue is whether or not abortion is murder.& (0)& I simply gave reasons why a woman might choose to abort and supported that.\\
\hline
\hline
\multicolumn {4}{|c|}{LAM\_LEX \cite{Vilares:17}}\\
\hline
& Viewpoint 1& & Viewpoint 2\\
\hline
(\textbf{-})& \textcolor{red}{abortion is NOT the only way to escape raising a child that would remind that person of something horrible}& (\textbf{+})& \textcolor{blue}{if a baby is raised by people not ready, or incapable of raising a baby, then that would ruin two lives.}\\
(\textbf{+})& \textcolor{blue}{I wouldn't want the burden of raising a child I can't raise}& (\textbf{+})& \textcolor{blue}{The fetus really is the mother's property naturally}\\
(0)&  a biological process is just another name for metabolism& (0)& Now this is fine as long as one is prepared for that stupid, implausible, far-fetched, unlikely, ludicrous scenario\\
(0)& The passage of scripture were Jesus deals with judging doesn't condemn judging nor forbid it& (0)& you are clearly showing that your level of knowledge in this area is based on merely your opinions and not facts.\\
(0)& your testes have cells which are animals& (0)& we must always remember how life is rarely divided into discreet units that are easily divided\\
\hline
\hline
\multicolumn {4}{|c|}{TAM \cite{Paul:10}}\\
\hline
& Viewpoint 1& & Viewpoint 2\\
\hline
(\textbf{-})& \textcolor{red}{I think that is wrong in the whole to take a life.}& (\textbf{+})& \textcolor{blue}{Or is the woman's period also murder because it also is killing the potential for a new human being?}\\
(\textbf{-})& \textcolor{red}{I think so it prevents a child from having a life.}& (\textbf{-})& \textcolor{red}{it maybe then could be considered illegal since you are killing a baby, not a fetus, so say the fetus develops into an actuall baby}\\
(\textbf{+})& \textcolor{blue}{Abortion is not murder because it is performed before a fetus has developed into a human person.}& (0)& In your scheme it would appear to be that there really is no such thing as the good or the wrong.\\
(0)& He will not obey us.& (0)& NO ONE! but God.\\
(0)& What does it have to do with the fact that it should be banned or not?& (0)& What right do you have to presume you know how someone will life and what quality of life the person might have?\\
\hline
\end{tabular}
\end{center}
\vspace{-45pt}
\end{table*}
For the following evaluations, we conduct a human annotation task with three annotators. The annotators are acquainted with the studied issues and the possible reasons conveyed by each side. They are given lists of mixed sentences generated by the models. They are asked to indicate the stance of each sentence when it contains any kind of persuasion, reasoning or argumentation from which they could easily infer the stance. Thus, if they label the sentence, the sentence is considered a relevant reason. The average Kappa agreement between the annotators is $0.66$. The final annotations correspond to the majority label. In the case of a conflict, 
we consider the sentence irrelevant.

We consider measuring the relevance by the ratio of the number of relevant sentences divided by the total number of the digest sentences.
Table \ref{RelAcc-table} contains the median relevance rate (Rel) over 5 runs of the models, on all datasets.  
PhAITV-based pipeline realizes very high relevance rates 
and outperforms its rivals, TAM and LAM\_LEX,
by a considerable margin on all datasets. Moreover, it 
beats its unigram counterpart AITV.
These results are also showcased in Table \ref{sample-output}'s examples. The ratio of sentences judged as reasons given to support a stance
is higher for PhAITV-based digest.  Interestingly, even the PhAITV's sentences judged as irrelevant are not off-topic, and may denote relevant argument facets like ``abortion is murder". 
Results confirm our hypothesis that phrasal facet argument leads to a better 
reasons' extraction.

All compared models generate sentences for each viewpoint. Given the human annotations, we consider assessing the viewpoint clustering of the relevant extracted sentences by two measures: the Clustering Accuracy and the Negative Predictive Value of pairs of clustered sentences(NPV).
NPV consider a pair of sentences as unit. 
It corresponds to the number of true stance opposed pairs in different clusters divided by the number of pairs formed by sentences in opposed clusters. 
A high NPV is an indicator of a good inter-clusters opposition i.e., a good contrast of sentences' viewpoints.
Table \ref{RelAcc-table} contains the median NPV and Accuracy values over 5 runs.
Both AITV and PhAITV-based frameworks achieve very encouraging NPV and accuracy results without any supervision. 
PhAITV outperforms significantly the competing contrastive summarization methods.
This confirms the hypothesis that leveraging the reply-interactions, in online debate, helps detect the viewpoints of posts and, hence, correctly cluster the reasons' viewpoints.
Table \ref{sample-output} shows a much better alignment, between the viewpoint clusters and the stance signs of reasons (+) or (-), for PhAITV comparing to competitors. The NPV and accuracy values of the sample digests are close to the median values reported in Table \ref{RelAcc-table}. The contrast also manifests when similar facets are discussed but by opposing viewpoints like in ``life begins at conception" against ``fetus before it can survive outside the mother's womb is not a person".
The results of PhAITV are not close yet to the PhAITV\textsubscript{view} 
where the true posts' viewpoint are given.
This suggests that the framework 
can achieve very accurate performances by enhancing viewpoint detection of posts.
\section{Conclusion}
\label{conclusion}

This work proposes an unsupervised framework for the detection, clustering, and displaying of the main sentential reasons conveyed by divergent viewpoints in contentious text from online debate forums. 
The reasons are extracted in a contrastive digest table. 
A pipeline approach is suggested based on a Phrase Mining module and a novel Phrase Author Interaction Topic-Viewpoint model. 
The evaluation 
of the approach
is based on three measures computed on the final digest: 
the informativeness, the relevance, and the accuracy of viewpoint clustering. The results on contentious issues 
from online debates 
show that our PhAITV-based pipeline outperforms state-of-the-art methods for all three criteria.
In this research, we dealt with contentious documents in online debate forums, which often enclose a high rate of rebuttal replies.
Other social media platforms, like Twitter, may not have rebuttals as common as in online debates. Moreover, a manual inspection of the digests suggests the need for improvement in the detection of semantically similar reasons and their hierarchical clustering.

\bibliography{cicling19}
\bibliographystyle{splncs04}
\end{document}